\documentclass[a4paper,fleqn]{cas-sc}

\usepackage[authoryear]{natbib}
\usepackage{soul}
\usepackage{multirow}
\def\tsc#1{\csdef{#1}{\textsc{\lowercase{#1}}\xspace}}
\tsc{WGM}
\tsc{QE}

\usepackage{subcaption}

\newcommand{\vecX}{\mathbf{x}}

\graphicspath{ {figs_rev1/} }
 
\usepackage{amsfonts} 
 
\def\tsc#1{\csdef{#1}{\textsc{\lowercase{#1}}\xspace}}
\tsc{WGM}
\tsc{QE}
\tsc{EP}
\tsc{PMS}
\tsc{BEC}
\tsc{DE}

\usepackage{lineno}

\begin{document}

\let\WriteBookmarks\relax
\def\floatpagepagefraction{1}
\def\textpagefraction{.001}
\shorttitle{Self-supervised similarity models based on well-logging data}
\shortauthors{S.Egorov et al.}

\title [mode = title]{Self-supervised similarity models based on well-logging data}

\author[a, b]{Sergey Egorov}
\credit{Software, Validation, Investigation, Writing - Original Draft, Writing - Review \& Editing, Visualization}

\author[a, b]{Narek Gevorgyan} 
\credit{Software, Validation, Writing - Original Draft}

\author[a]{Alexey Zaytsev}
\credit{Conceptualization, Methodology, Formal analysis, Supervision, Funding acquisition, Writing - Review \& Editing}

\address[a]{Skolkovo Institute of Science and Technology, Moscow, Russia}
\address[b]{These authors contributed equally to this work.}

\begin{abstract}
Adopting data-based approaches leads to model improvement in numerous Oil\&Gas logging data processing problems. These improvements become even more sound due to new capabilities provided by deep learning. However, usage of deep learning is limited to areas where researchers possess large amounts of high-quality data. We present an approach that provides universal data representations suitable for solutions to different problems for different oil fields with little additional data. Our approach relies on the self-supervised methodology for sequential logging data for intervals from well, so it also doesn't require labelled data from the start. For validation purposes of the received representations, we consider classification and clusterization problems. We as well consider the transfer learning scenario. We found out that using the variational autoencoder leads to the most reliable and accurate models. approach  We also found that a researcher only needs a tiny separate data set for the target oil field to solve a specific problem on top of universal representations.
\end{abstract}


\maketitle

\printcredits

\section{Introduction}
\label{sec:introduction}

Recently many deep learning approaches have been applied to time series analysis 
for  Oil\&Gas data ~\cite{Development_of_Deep, Towards_better}.
Most of them leverage a large amount of labelled data for training. However, often it is impossible or at least complex to get the corresponding dataset: the data labelling is both resource-consuming, and subjective  ~\cite{fredriksson2020data}. 

A sound alternative is to use unlabeled data for model training.
In particular, a common approach in computer vision and natural language processing consists of two steps.
The first is to learn a representation model, and the second is to train a small model on top of the representation.
The representation learning model is designed to automatically extract relevant features with no requirement of extensive annotated data making manual feature engineering superfluous. In the context of this work, the focus is on unsupervised methods that learn a compact representation without knowledge of the target variable and thus only based on the multivariate time series.
Typically, a sizeable data-hungry foundation model aggregates information from vast amounts of unlabeled data.
If the representation model is universal enough, we require little tuning of the representation model and an adjustment on top of it.
This adjustment has few parameters, thus requiring a small amount of data to tune it and solve the problem.

The work focuses on Self-supervised learning, one of the most popular methods to overcome the problems we face with a supervised approach. The idea is based on pseudo-labels - representations created by the learning algorithm and later used in supervised learning. The methods could be classified into generative and contrastive learning-based approaches. In ~\cite{romanenkova2022similarity, rogulina2022similarity}, authors used contrastive learning-based models to derive a latent space such that intervals having close characteristics would be located close to each other, while different intervals are placed as far as from each other as possible. Meanwhile, in ~\cite{gan}, GANs were used to resolve the issues related to incompleteness and noisiness of well-logging data. In this work, we focus on autoencoder-based approaches because, in contrast to GANs, these generally provide a better representation of the training data ~\cite{liu2021self}. Also, we learn Deep Autoregressive models, which produce embedding of the wells capable of predicting the following timestamp values and compare the representations with ones derived using the contrastive loss technique.

We opt for two strategies to validate the representations generated by the considered self-supervised models. The first strategy refers to using the same data. The second strategy is to validate the representations by estimating the transfer learning capabilities of the derived embeddings using new data. We focus on transfer learning since this family of methods is used in almost every deep learning model when the target dataset does not contain enough labelled data. Despite its recent success in computer vision, transfer learning has rarely been applied to deep learning models for time series data. One of the reasons for this absence is probably the lack of one extensive general-purpose dataset similar to ImageNet or OpenImages but for time series. Furthermore, it is only recently has been proven that deep learning works well for time series classification, and there is still much to be explored in building deep neural networks for mining time-series data.

 There is little research in the Oil$\&$Gas industry related to using representations and parameters of neural networks. For example, we came across the paper~\cite{liu2020well}, which reports on applying transfer learning of deep neural networks to well-logging data. However, most studies consider only fine-tuning as model completion or approach based on expert markup and data preprocessing, for example, the similarity between the source data task and the target task. We focus on approaches encouraging the similarity of representations or parameters of deep neural networks. Transfer learning is attracting more and more attention, including in the Oil$\&$Gas industry, primarily due to the lack of need for large volumes of training data and proper model selection. Training allows using a pre-trained model with customized behaviour patterns that are transferred to a target task. Nevertheless, the simple use of additional training ~\cite{Evaluation_of_transfer}, or fine-tuning, does not significantly increase the metrics of the mode. For example, in ~\cite{Reservoir_Production}, the use of fine-tuning with the minimization of the MSE functionality can improve the quality by only 3 percent. A fine-tuning application is often used - selective freezing of neural network layers ~\cite{A_Deep_Transfer}. Another common method is data preprocessing ~\cite{liu2020well} based on shuffling ~\cite{Towards_better}, bootstrapping or population-based evolutionary algorithms ~\cite{Dynamic_production}.

We propose a new approach in this area to transfer learning, using the similarity of the parameters of the source and target task models based on the $l_2$ regularizer, as well as the similarity of the representations of the hidden layers of the RNNs.
The models constructed via the proposed approach can transfer well to new oil fields and require little labelled data to provide high quality for them.
In many cases, the model can be used as it is without fine-tuning.
Moreover, it doesn't suffer from catastrophic forgetting, maintaining memory about all presented training data.

We structure the paper in the following way to describe the results:
\begin{itemize}
    \item In Section~\ref{sec:methods}, we consider methods that we present for self-supervised learning. We also discuss transfer learning applications to time series data and, in particular, time series from the Oil\&Gas industry.
    \item In Section~\ref{sec:data}, we provide an overview of the data and preprocessing for our experiments.
    \item In Section~\ref{sec:results}, we present the results of our numerical experiments that examine the quality of the methods considered earlier.
    \item Finally, Section~\ref{sec:conclusion} presents the overall conclusions of the work.
    \item Additional experiments on the Transfer learning capabilities of similarity models, as well as the selection of hyperparameters, are given in Appendix~\ref{sec:transfer_app}.
\end{itemize}





\section{Methods}
\label{sec:methods}

\subsection{Self-supervised learning}

The most popular and ideal setting in the machine learning world is training models with labelled objects. We have a target value for each object that our model should predict after being trained with available data. In the real world, most of the available data are unlabeled~\cite{romanenkova2022similarity}. 
For well drilling logging data, we do not manage to determine the corresponding formation~\cite{rogulina2022similarity} or a rock type~\cite{klyuchnikov2019data} for a specific well without supervision from an expert. However, we can utilize these data to construct representations suitable for many so-called downstream tasks. Learning from unlabeled data to get representations generated by an encoder is called self-supervised learning. Two main types of self-supervised training are generative (or autoencoder-based) and contrastive approaches~\cite{liu2021self}.

The idea of the generative approach is to learn two models: an encoder and a decoder. The encoder process a
description of an object and produces an embedding that acts as a low-dimensional representation of the original
input. Meanwhile, the decoder transforms features extracted from the previous step and produces a reconstruction of the
initial description of the object, in other words, input data of the encoder.
We train both models in an
end-to-end learning framework with a loss that forces a successive application to the initial input that produces an
output close to the input itself. Thus, our embedding holds most of the information about the processed object.

For contrastive learning, we assume that only a single encoder model exists with maybe a simple discriminator
on top of it. We train the encoder so that representations of similar objects are close in the embedding space while
representations of different objects are far away.
The article compares generative and contrastive approaches to self-supervised learning and how they work for
different downstream problems. Moreover, we will introduce an autoregressive model with its performance on analogous
tasks.

\subsubsection{Autoencoders}
\label{subsec:aggregation}

An autoencoder ~\cite{liu2021self} is one of the possible architectures of neural networks that are supposed to have its output as similar to the input as possible. 
To ensure this property first part of the intermediate layers of the network compresses the input into a lower-dimensional representation and the remaining part of the layers reconstruct the output from this representation. 
The compressed result derived after training the model acts as an embedding of the input in a latent space. The first part compressing input and producing compressed representation is called the encoder. The decoder reconstructs input using previously derived embedding.

\begin{figure}[!hb]
    \centering
    \includegraphics[width=0.7\linewidth]{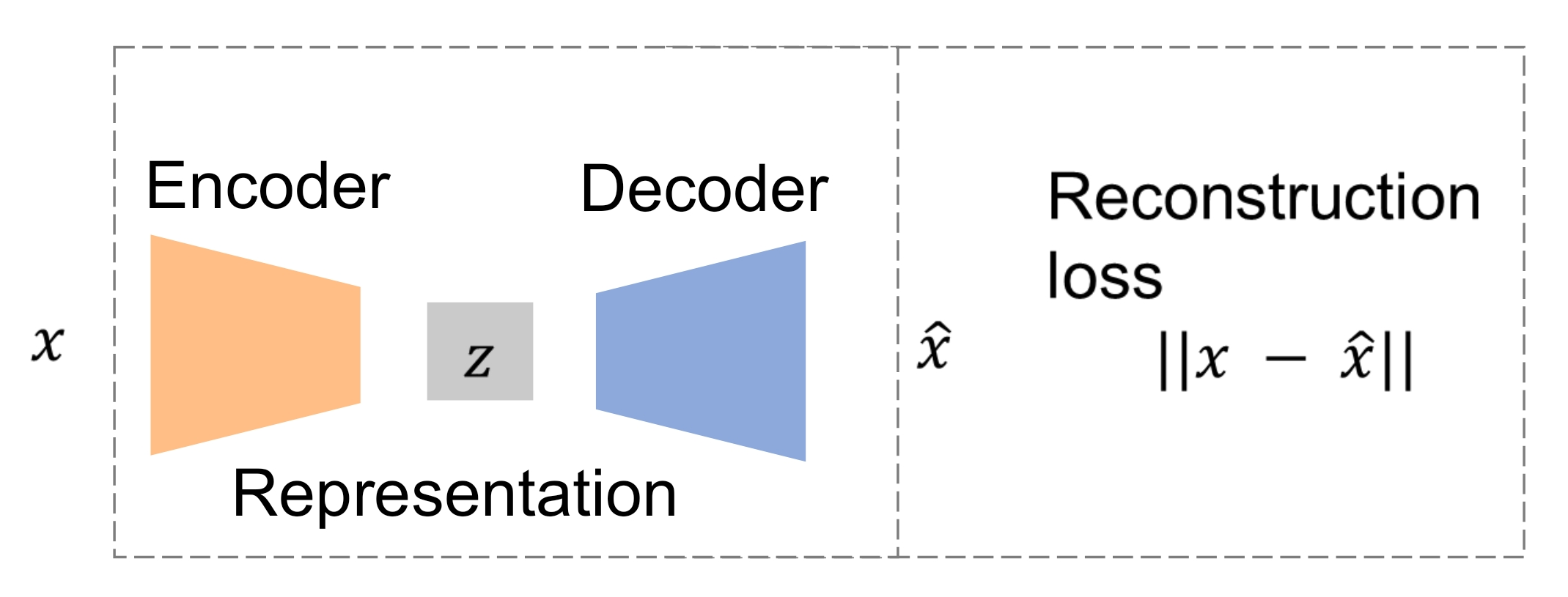}
    \caption{Autoencoder model structre}
    \label{fig:basic_ae}
\end{figure}

The structure of the encoder is shown in Figure ~\ref{fig:basic_ae}.

To overcome the problem of lack of labeled data we can frame the learning task as a supervised problem considering the reconstruction $\hat{x}$ of the original data $x$ as an output of the model. To train the network we will optimize reconstruction loss $\mathcal{L}(x, \hat{x})$, that estimates how well the reconstructed time series approximate original data.

We can take an unlabeled dataset $x$ and frame it as a supervised learning problem tasked with outputting a reconstruction $\hat{x}$ of the original input. This network can be trained by minimizing the reconstruction error $\mathcal{L}(x, \hat{x})$, which measures the differences between our original input and the consequent reconstruction. The mode essential element of the autoencoder architecture is a bottleneck, a hidden layer with the lowest dimension that is responsible for producing the embedding. It is used both to get a lower-dimensional vector describing the input and to preserve the network from memorizing the processed data.
\begin{equation} 
\label{eqn:ae_loss}
    L_{AE} = ~\parallel x - \hat{x}\parallel^2_2
\end{equation}


The loss function of an autoencoder is designed in such a way that the network is
capable of yielding matching output for each data object and still does not overfit against training data by memorizing it. To fulfill the purpose, the loss usually consists of two components: the first term, the reconstruction loss, is sensitive to the inputs, and the second regularizer term prevents memorization. Furthermore, it is common to use scaling parameters for the second term to control the trade-off between the two goals of the model.

\subsubsection{Variational autoencoders}

The distribution of the representations in the latent space we derive using autoencoder is arbitrary. 
This issue leads to degraded quality of embeddings and reconstructions.
Variational autoencoders ~\cite{variationalae} deal with this problem by introducing more constraints to the embeddings.

A variational autoencoder is a stochastic approach that maps each input object to a probability distribution, whereas the traditional autoencoder learns a deterministic mapping from the input space to the latent space. To elaborate further, the model learns a function from the bottleneck hidden layer to the layers that represent mean $\mu(x)$ and variance $\sigma(x)$ of the corresponding diagonal Gaussian distribution.

Despite the changes in the architecture, the variational autoencoder still could output embeddings having significantly different mean and variance values. As a consequence, that nevertheless creates non-continuous latent space. To solve this issue, the autoencoder suggests an auxiliary loss that places an additional constraint on the mentioned Gaussian distributions $p(z|x)$, forcing them to stay as close to the standard normal distributions $\mathcal{N}(0, 1)$ as possible.

The auxiliary loss should be designed with the intention to penalize distributions $p(z|x)$ for differing from the predefined normal distribution $\mathcal{N}(\mu, \sigma)$. At the same time, the form of the appropriate loss should have properties to be used with gradient descent. One of the most common measures that satisfy the mentioned requirements is KL divergence between $p(z|x)$ and $\mathcal{N}(\mu, \sigma)$. In the particular case when $\mu = 0$ and $\sigma = 1$, we derive a loss ~\ref{eqn:kl_divergence} used in the variational autoencoder.
\begin{equation} 
\label{eqn:kl_divergence}
    KL(\mathcal{N}(\mu, \sigma) \parallel  \mathcal{N}(0, 1)) = \frac{1}{2}\sum_{x \in X} (\sigma^2 + \mu^2 - \log (\sigma^2) - 1).
\end{equation}

\subsubsection{Adversarial autoencoders}

Despite the popularity of variational autoencoders, applying the approach is not simple due to the KL divergence term used in the loss function. The first complication is that defining a closed-form solution for most distributions is unfeasible. In addition, it is not always clear how to apply the propagation of derivatives through discrete nodes. As we have seen before, additional hidden layers could be used for this purpose in the case of variational autoencoders, although this makes the training procedure less efficient.

Adversarial autoencoders avoid using the KL divergence altogether by using adversarial learning~\cite{makhzani2015adversarial}. In this architecture, a new network is trained to predict discriminatively whether a sample comes from the hidden code of the autoencoder or the prior distribution $p(z)$ determined by the user. The encoder's loss is now composed of the reconstruction loss plus the loss given by the discriminator network.



On the adversarial regularization part, the discriminator receives $z$ distributed as $q(z|x)$ and $z'$ sampled from the true prior $p(z)$ and assigns a probability to each of coming from $p(z)$. The loss incurred is backpropagated through the discriminator to update its weights. Then the process is repeated, and the generator updates its parameters.

We can now use the loss incurred by the generator of the adversarial network, which is the encoder of the autoencoder, instead of the KL divergence to learn how to produce samples according to the distribution $p(z)$. This modification allows us to use a broader set of distributions as priors for the latent code.
For example, one can use multimodal distributions.

The loss of the discriminator is:
\[
L_{D} = - \frac{1}{m} \sum_{k=1}^{m} log(D(z')) + log(1 - D(z)),
\]
where $m$ is the batch size, $z$ is generated by the encoder, and $z'$ is a sample from the true prior.

For the adversarial generator we have:
\[
L_{G} = - \frac{1}{m} \sum_{k=1}^{m} log(D(z)) 
\]

Considering the equations, we can see that the loss defined this way will force the discriminator to be able to recognize fake samples while will push the generator to fool the discriminator.

\subsubsection{Autoregressive methods}
\label{subsec:aggregation}

Autoregression is predicting a future outcome of a sequence from the previously observed outcomes of that sequence. More formally: 
\[
\vecX_{t+1} = \sum_{i=0}^{m} \delta_{i} \vecX_{t-i} + c.
\]
It looks a lot like linear regression, and that is because, mathematically, it effectively is. It uses the previous $m$ terms to predict the next term, where $m$ is a constant called a lag or a receptive field. 

DARNs ~\cite{darns} (Deep AutoRegressive Networks) are generative sequential models and are therefore often compared to other generative networks like GANs or VAEs; however, they are also sequence models and show promise in traditional sequence challenges like language processing and audio generation.

Technically, any network that uses previous data from a sequence to predict a future value in that sequence could be autoregressive, but when in the context of deep learning, autoregression almost always refers to the relation of prior outputs as inputs as opposed to recurrent models which take a set amount of predefined input. To clarify, outputs are fed back into the model as input, making the model autoregressive. Usually, the implementation ends up being a convolutional layer or series of convolutional layers with autoregressive connections.


Most deep learning autoregressive networks formally define a probabilistic model that connects past and future observations:
\[
\hat{f}(\vecX_0, \ldots, \vecX_t) = \prod_{t=1}^{T} P(\vecX_{t+1}| \vecX_{t}, \ldots, \vecX_0).
\]

So by learning an autoregressive relationship, we actually model the probability distribution of 
$\vecX$. It is pretty simple to implement, and, moreover, this is where we introduce additional limitations on the autoencoder. All we need to do is restrict autoencoder connections so that the node for predicting $\vecX_t$ is only connected to the inputs  $\vecX_{1:t-1}$. 

\subsubsection{Contrastive learning}
\label{subsec:contrastive}

The goal of contrastive representation learning is to learn such an embedding space in which similar sample pairs stay close to each other while dissimilar ones are far apart. Contrastive learning can be applied to both supervised and unsupervised settings. When working with unsupervised data, contrastive learning is one of the most powerful approaches in self-supervised learning. As shown in~\cite{romanenkova2022similarity}, the approach could be efficient, particularly for representation learning of time series.

Contrastive loss is one of the earliest training objectives used for deep metric learning in a contrastive fashion. Given a list of input samples ${\vecX_i}$, each has a corresponding label $y_i$. We would like to learn a function $f: \mathcal{X} \rightarrow \mathbb{R}^d$ that encodes $\vecX_i$ into an embedding vector such that examples from the same class have similar embeddings and samples from different classes have very different ones. Thus, contrastive loss takes a pair of inputs $(\vecX_i, \vecX_j)$ and minimizes the embedding distance when they are from the same class but maximizes the distance otherwise.
\[
 \mathcal{L}(\vecX_i, \vecX_j) = [y_i = y_j]\parallel{f(\vecX_i) - f(\vecX_j)}\parallel_{2}^{2} + \mathbb{1}[y_i \neq y_j] \max{(0, \epsilon - \parallel{f(\vecX_i) - f(\vecX_j)}\parallel_{2}^{2})} 
\]

where $\epsilon$ is a hyperparameter, defining the lower bound distance between samples of different classes.

Triplet loss was originally used to learn face recognition of the same person at different poses and angles. Given one anchor input $\vecX$ we select one positive sample  $\vecX^{+}$ and one negative $\vecX^{-}$, meaning that $\vecX^{+}$ and $\vecX$ belong to the same class and $\vecX^{-}$ is sampled from another different class. Triplet loss learns to minimize the distance between the anchor $\vecX$ and positive  $\vecX^{+}$ and maximize the distance between the anchor  $\vecX$ and negative $\vecX^{-}$ at the same time with the following equation:
\[
 \mathcal{L}(\vecX, \vecX^{+}, \vecX^{-}) = \sum_{\vecX \in \mathcal{X}} \max{(0, \parallel{f(\vecX) - f(\vecX^{+})}\parallel_{2}^{2} - \parallel{f(\vecX) - f(\vecX^{-})}\parallel_{2}^{2} + \epsilon)}
\]

where the margin parameter $\epsilon$ is configured as the minimum offset between distances of similar vs dissimilar pairs. It is crucial to select $\vecX^{-}$  challenging to truly improve the model.

\subsection{Self-supervised models validation}

Evaluating representations obtained with self-supervised methods is not that straightforward due to the absence of apparent target values. The standard routine, in this case, is to derive some labels based on the given data and then use them in a supervised manner. Nevertheless, supervised evaluation is not always manageable, especially when applied to the time series. Therefore, we used cluster-based and transfer learning-
based approaches to assess previously derived representations of the series.

\subsubsection{Clustering-based evaluation}

A supplementary clustering task could be useful to compare representation learning methods. The main challenge of the approach is to choose a reasonable strategy to group the data and later predict them based on the corresponding embeddings of the time series.
Making use of several categorization techniques gives more robust statistics.

More precisely, we cluster the representations using one of the following alogorithms:

\begin{enumerate}
    \item K-Means Clustering
    \item Clustering using Gaussian Mixture Models
    \item Agglomerative Hierarchical Clustering
\end{enumerate}

For clustering we utilize three main metrics~\cite{romanenkova2022similarity}:

\begin{enumerate}
    \item The Adjusted Rand Index (ARI) is Rand Index corrected for the chance of random label assignments:
\[
  \textbf{ARI} = \frac{RI- \mathbb{E}[RI]}{\max{(RI)} - \mathbb{E}[RI]}
\]

If we denote $\textit{n}$ the total number of objects from the validation sample, $\textit{a}$ the number of pairs in the same cluster and have the same label, $\textit{b}$ the number of pairs that are in different clusters and have different labels, then:
\[
  \textbf{RI} = \frac{a + b}{C^n_2}
\]

    \item The Adjusted Mutual Information (AMI) is Mutual Information adjusted for the chance of random label assignments::
\[
  \textbf{AMI} = \frac{MI- \mathbb{E}[MI]}{mean(H(U),H(V)) - \mathbb{E}[MI]}
\]

Assuming, two label assignments $U$ and $V$, we denote $P_i = \frac{|U_j|}{n}$ and $P_j' = \frac{|V_j|}{n}$. Then the entropy of U is 
\[
  H(U) = - \sum_{i=1}^{|U|} P_i \log{P_i}
\]

Similarly we define for V. If we denote $P_{ij} = \frac{|U_i \cap V_i|}{n}$, then mutual information(MI) between U and V:
\[
  MI(U, V) = P_{ij}\log\frac{P_{ij}}{P_i P_j'}
\]

    \item The V-measure or Normalised Mutual Information is the harmonic average between completeness and homogeneity, which can be written in the following form:
\[
\textbf{V-measure} = \frac{2MI(U, V)}{H(U) + H(V)}
\]
\end{enumerate}

\subsubsection{Transfer Learning}
\label{subsec:aggregation}

For the purpose of validation, we also use concepts of the transfer learning approach. 
More precisely, in this section, we try to figure out how well a model can generalize from a source to a target domain and later use the quality of the transfer metric to compare considered models.

Transfer learning is a machine learning paradigm aiming at transferring the knowledge obtained in a source task to a new target task. While a source domain and a target domain differ, they share common features, and thus, an encoder trained using the source data is an excellent initial guess for an encoder tailored to the target data. Here we will use a neural network-based transfer learning approach - it refers to reusing the structure and connections parameters of the network pretrained in the source domain. The basic idea is to use the model with parameters $w^{0}$ estimated at the source task and transfer them to get good quality for a target task model with parameters $w$. This task is easier than learning from scratch since the model can be trained on significantly less data. In addition, the use of ready-made representations significantly accelerates machine learning. It solves the problem of data dependence - one of the most severe problems in traditional deep learning, which refers to the need to train a network on large amounts of data to understand latent patterns.

\paragraph{Fine-Tuning.}

A simple yet effective approach to obtaining high-quality deep learning models is performing fine-tuning weights. 
In such practices, a deep neural network is first trained using a large source dataset. The weights of a network are then fine-tuned using the data from the target application domain via minimization of $\sum_{i=1}^{n}L(z(\vecX_i, w), y_i)$.

Fine-tuning is a specific approach to performing transfer learning in deep learning. The weights $w^{0}$ pretrained by the source dataset with a sufficiently large number of instances usually provide a better initialization for the target task than random initialization. 

In a typical fine-tuning approach, weights in lower layers are fixed, and weights in upper layers are tuned using data from the target domain. 

Fine-tuning can lead to the situation in the case that the parameters of the target model may be driven far away from initial values, which also causes overfitting in transfer learning scenarios.
So, researchers often impose additional constraints to avoid overfitting issues and improve the quality of the model.

\paragraph{}

Therefore, effective work of vanilla transfer learning methods such as fine-tuning still requires an adequate amount of training data for target task. Failure to comply with this condition entails two problems - catastrophic forgetting and negative transfer. The first means the tendency of models to lose previously learned patterns that may contain information relevant to the target task, eventually leading to overfitting.
To deal with it, there are several approaches with the most common one being regularization.
We consider the state-of-the-art regularization described in ~\cite{l2}.
There the authors investigate regularization schemes to accelerate deep transfer learning while preventing fine-tuning from over-fitting. 
Their work showed that a simple $L^2$-norm regularization on top
of the “Starting Point as a Reference” optimization can significantly outperform a wide range of
regularization-based deep transfer learning mechanisms, such as the standard $L^2$-norm regularization. 

With such
regularization, the transfer learning problem can be reduced to the similarity learning problem in the following way.

Let’s denote the dataset for the target task as $D = \{(\vecX_i, y_i)\}_{i = 1}^n$, where totally $n$ tuples are offered, and each tuple refers to the input sample and its label in the dataset. With such
regularization, the transfer learning problem can be reduced to a similarity learning problem as follows:
\[
\sum_{i=1}^{n}L(z(\vecX_i, w), y_i) + \lambda \Omega(w, w^{0}, D) \rightarrow \min_{w}.
\]
The first term $\sum_{i=1}^{n}L(z(\vecX_i, w), y_i)$ refers to the empirical loss of data fitting quality for  model $z$, while the second term is a general form of a regularization loss $\Omega(w, w^{0}, D)$. 
The parameter $\lambda > 0$ balances the empirical loss and the regularization loss.
The vector of parameters $w^{0}$ refers to the parameters obtained via training at the source task.

\paragraph{$L^{2}-sp$ regularization}

Using these initial parameters $w^{0}$ as the reference in the $L^2$ penalty, we get the following regularizer $\Omega(w, w^{0})$:
\[
\Omega(w, w^{0}) = \alpha \mid\mid w - w^{0}\mid\mid_{2}^2,
\]
$\alpha$ refers to the weight of $l_2$ regularlizer in the total regularizer loss.

\paragraph{Delta: Deep Learning Transfer using Feature Map}

Delta~\cite{li2019delta} intends to incorporate a new regularizer $\Omega(w, w^{0}, \vecX)$.  Given a pre-trained parameter $\omega^{0}$ and any input $\vecX$, the regularizer $\Omega(w, w^{0}, \vecX)$ measures the distance between the behaviors of the target network with parameter $\omega$ and the source one based on $\omega^0$. Thus, the regularizer takes the following form:
\[
\Omega(w, w^{0}, D) = \sum_{i=1}^{n}\Omega(\omega,\omega^{0},\vecX_i,y_i)
\]

where $\sum_{i=1}^{n}\Omega(\omega,\omega^{0},\vecX_i,y_i,z)$ characterizes the aggregated difference between the source and target network over the whole training dataset.

To regularize the behavior of the networks, Delta considers the distance between the outer layer outputs of the two networks. Given an input $\vecX_i$, each filter generates a feature map. Thus, Delta characterizes the outer layer output of the network model $z$ based on input $\vecX_i$ and parameter $\omega$ using a set of feature maps, such as $FM_{j}(z, \omega, x_i)$ for $L$ layers in the network. In this way, the behavioral regularizer is defined as:
\[\Omega(\omega,\omega^{0},\vecX_i,y_i,z) = \sum_{j=1}^{L}{||FM_{j}(z, \omega, x_i) - FM_{j}(z, \omega, x_i)||_2^2}
\]

\paragraph{Batch Spectral Shrinkage}

BSS ~\cite{NIPS2019_8466} is a new regularization approach, detailed procedures are as follows:

\textmd{1) Constructing a feature matrix $\textbf{F}$ from a batch size $b$ of feature vectors $\vecX$}

\textup{\textmd{2) Applying ${S}{V}\!{D}$ to compute all singular values of $\textbf{F}$ as Equation $$F=U{\Sigma}V^T$$}} 

\textup{\textmd{3) Penalizing the smallest $k$ singular values $[\sigma_1, \sigma_2,..., \sigma_b]$ in the diagonal of}}

\textup{\textmd{singular value matrix $\Sigma$ to mitigate negative transfer:}}

$$L_{bss}(F) = \eta\sum_{i=1}^{k}{\sigma_{-i}^2}$$

After this, BSS embedded into existing fine-tuning regularizer scenarios can be formulated as:

\[
\Omega(\omega) + L_{bss}(F).
\]

\subsection{Conclusions}
\label{subsec:tl_conclusion}

We will combine above described transfer learning approaches to evaluate the generalization capability of the representation learning models. More precisely, we consider four self-supervised methods with hyperparameters:

\begin{enumerate}
    \item Triplet loss-based method 
    \item Variational \& Adversarial Autoencoders
    \item Autoregressive Model
    \item Siamese
\end{enumerate}

For each of the given methods, we choose the best model based on their performance on the source task. 

Later, to add one more layer of comparison of different models, we consider four transfer learning scenarios:

\begin{enumerate}
    \item Base case: train source model on target domain without any additional restrictions 
    \item Fine-tuning: train source model on target domain while keeping some of the layers fixed, i.e. weights of the corresponding layers stay the same as for source model
    \item $L^2 - sp$ regularization: train source model on target domain with an additional loss on weights of the model that force them to change as little as possible
    \item Delta: train source model on target domain with an additional loss on some layers outputs that force them to change as little as possible
\end{enumerate}

We also present the BSS add-on in Appendix~\ref{sec:hyperparameters_self_supervised} as an application to avoid the negative transfer.

In the process, we compare different approaches via various ablation studies and supplementary problem statements.
These experiments help to better shape our results and identify the scope of applicability of developed models.

\section{Data overview}
\label{sec:data}
We use the open access dataset from Taranaki Basin provided by the New Zealand Petroleum \& Minerals Online Exploration Database \cite{web:newzeland1}, the Petlab \cite{web:newzeland2} and also the dataset from the Norwegian offshore yield by Norwegian Petroleum Directorate \cite{web:norwegian}. 

In general, in preprocessing, we adopt ideas from~\cite{romanenkova2022similarity}, and choose the following characteristics of wells as features: sonic log (DTC), gamma-ray (GR), density log (DENS), porosity inferred from density log (DRHO). As a source formation, we choose Urenui, which consists of 31 wells after preprocessing. For the purpose of transfer learning, we use data from Moki, Matemateaonga and Mohakatino formations with a total of 104 wells. The Norwegian data is represented by the Utsira formation, which consists of 12 wells. All the selected formations have an average length of near 3600 feet.

\subsection{Sampling}

We sample from the wells pairs of intervals of a fixed length of $l =$ 100 measurements with a depth difference between two consecutive measurements is one foot for our experiments. With the average length of wells equal to 3600 feet, such interval length allows us to generate a sufficient number of samples for training while maintaining the structure of the well and, consequently, patterns of behavior in the resulting interval. Thus, we pair intervals of length 100 and assign a target value to the pair - 1 if the intervals are from the same well (or the distance between them does not exceed the specified parameter for Close Well Linking problem, more detailed in Section ~\ref{sec:transfer_app}), and the value 0 otherwise. As a result, the training sample has the following form: $\{(\{x^i_1, x^i_2, x^i_3, x^i_4\}^j, y^j \in \{0, 1\})\}^N_{j=1}$, where N is the size of training dataset and $x^i_1, x^i_2, x^i_3, x^i_4$ is DRHO, DENS, GR and DTC features accordingly.
\section{Results}
\label{sec:results}

In this section, we describe an experimental evaluation of the proposed ideas. 
The self-supervised architectures and corresponding models are specified in more detail, followed by their performance comparison. 

\subsection{Validation approaches}

The amount of labelled data is insufficient, and labels are expert-based and, thus, subjective. 
However, we use supervised data with the derived target labels to evaluate the representations obtained using each proposed model. 
We adopt two approaches to valuation via labelled data.
\textit{The first approach} is to compare the adequacy of the spatial distribution for embeddings obtained via our encoder models. 
Clustering the encoder's embeddings provides cluster labels that we compare with true ones. 
We utilize state-of-the-art approaches for clustering that explore well the spatial structure of the embedding space, including KMeans, Gaussian Mixture Model, and Agglomerative Clustering~\cite{rokach2005clustering} available in a popular package python sklearn~\cite{scikitlearn}. 
\textit{The second approach} is the transition from self-supervised learning to the supervised task using solely obtained embeddings. As such a benchmark, we propose the task of the rock type prediction. On the one hand, our embeddings should contain information about the type of rock for an interval. On the other hand, this information is tied to depth and not to a particular well, while no explicit lithotype is available to a model during training.

\subsection{Model selection and training}

Initially, we considered three main self-supervised methods suitable for processing time series: contrastive methods~\cite{romanenkova2022similarity}, autoencoders~\cite{makhzani2015adversarial} and autoregressive approaches~\cite{radford2019language}.

To make the embeddings of time series more robust and increase the range of applicability of obtained representations, we use training data augmentations techniques and regularizer for the loss function. As data augmentation techniques , we choose two popular approaches. One of the approaches is based on the same idea as a denoising autoencoder. That is, for each of the input time series Gaussian noise is applied. Also, we could mask some parts of the input sequence with a predefined probability. These augmentations are suitable for usage with deep learning models ~\cite{masked_ae_cv}, ~\cite{masked_ae_de} and have been considered for oil drilling logging data~\cite{ae_for_well_logging}.

For the same reason, we combine the main model loss with an auxiliary loss of the supplementary task. The classification loss defined on the task, where we predict labels of the intervals that are assigned to the wells based on their coordinates, could be used with all the previously mentioned approaches. In other words, the representation learning model is constrained such way to incorporate topographical information of the interval. Another regularization technique is based on the autoregressive loss, which is simply an error of the model while predicting the time series values for the next n steps based on the interval log.

\subsection{Clustering validation}

\subsubsection{Base models}

We compare main contrastive and autoencoder-based models with all the possible modifications stated before. Since the formation labels for all the intervals in the training and validation datasets are obtained by the expert, we use them as a target for the clustering validation task. 

The results for two clusterization quality metrics, ARI and AMI, are in Table \ref{tab:ablation_ae}. The model settings for the experiment are given in the Table \ref{tab:transfer_settings}.
The best results are achieved on embeddings obtained by a Variational Autoencoder with additional classifier regularization.
In this specific case, Variational Autoencoder with given specifications significantly outperforms the results of our previous baseline contrastive model given in~\cite{romanenkova2022similarity}. 

\begin{table}[!h]
\centering
    \caption{Matching of true well labelling and labelling obtained via clusterization of embeddings for different representations. The best values are in \textbf{bold}. The second best values are \underline{underlined}.}
    \begin{tabular}{lllcc} 
    \hline
    Main loss & Regularization & Architecture & ARI & AMI\\ 
    \hline
    Triplet & - & LSTM & 0.08 & 0.20 \\
    Autoencoder & noise & LSTM & 0.09 & 0.20 \\
    Autoencoder & mask & LSTM & 0.13 & 0.21 \\
    Autoencoder & noise + mask & LSTM &  0.06 & 0.16 \\
    VAE & - & LSTM & \underline{0.24} & 0.34 \\
    VAE & classification loss reg. & LSTM &  $\mathbf{0.35}$ & \underline{0.35} \\
    VAE & classification loss reg. & CNN & 0.19 & $\mathbf{0.39}$ \\
    \hline
    \end{tabular}%
    \label{tab:ablation_ae}
\end{table}

\subsubsection{Best models for each clustering task}

Our main purpose is to obtain the self-supervised model that will define for each interval a corresponding representation that could be used in other tasks, which means the embeddings should have high generalization capabilities. To evaluate models in this domain, we consider the transfer learning approach in \ref{subsec:tl_results} and try to obtain other patterns of the embeddings clusters.

Here, we compare our embedding space with those provided by expert target labels. We devised the validation approach based on three labels for the obtained clusterization:
\begin{itemize}
    \item Geographical: label of the interval derived from clustering wells based on well coordinates
    \item Formation: expert-based formation label of the interval
    \item Formation\&class: expert-based (formation, class) pair
\end{itemize}

Our space of models here also includes Adversarial Autoencoder and Autoregressive approaches. 

For autoregression, we tried several values for the number of predictions; the resulting model, showing the best performance for all the clustering tasks, has LSTM as an encoder, classification auxiliary loss and three future values to predict.

The best models among considered for geographical labels, formation labels and (formation, class) pair labels are presented in Table~\ref{tab:geogr_target_clusterization}.
We select one leader for each model type: contrastive, autoencoder, and autoregression.

These experiments show that for all three clustering tasks, Variational Autoencoder (VAE) or Adversarial Autoencoder (AAE) appeared as the best representation models. If we take into account that VAE is a special case of  AAE, we conclude that AAE is the best model so far that could be further improved due to the large space of hyperparameters responsible for the generation of the embedding space.

\begin{table}[h!]
    \centering
    \caption{Three best models for different true target values: geographical, formation and pairs of formation and class pair. The best values are in \textbf{bold}. The second best values are \underline{underlined}.}
    \begin{tabular}{lllcccc}
    \hline
    Main loss & Regularization & Architecture & ARI & AMI  \\
    \hline
    \multicolumn{5}{c}{Geographical} \\
    \hline
    Triplet & - & LSTM & \underline{0.14} & \underline{0.26} \\
    VAE     & Classification loss reg & LSTM & $\mathbf{0.35}$ & $\mathbf{0.35}$ \\
    AR-4  & Classification loss reg + mask + noise & CNN & 0.07 & 0.22 \\
    \hline
    \multicolumn{5}{c}{Formation} \\
    \hline
    Triplet & - & LSTM & \underline{0.08} & \underline{0.20} \\
    VAE & Classification loss reg + mask & LSTM & $\mathbf{0.24}$ & $\mathbf{0.33}$\\
    AR-2 & Classification loss reg + mask + noise & CNN & 0.05 & 0.09 \\
    \hline
    \multicolumn{5}{c}{Formation and class} \\
    \hline
    Triplet & - & LSTM & 0.06 & 0.30 \\
    AAE & Autoregression loss reg & CNN & \underline{0.14} & \underline{0.38} \\
    AR-4  & Classification loss reg + mask + noise & CNN & $\mathbf{0.15}$ & $\mathbf{0.40}$ \\
    \hline
    \end{tabular}
    \label{tab:geogr_target_clusterization}
\end{table}

\subsection{Transfer learning via different self-supervised models}
\label{subsec:tl_results}

We consider the transfer of similarity machine learning models trained on Urenui formation from the New Zealand data.
As the target data, we consider Moki, Matemateaonga and Mohakatino formations from the New Zealand data as well as the Utsira formation from open Norway data.
Additional experiments for this section, in particular, hyperparameters selections, are in Appendix~\ref{sec:hyperparameters_self_supervised}.  

We compare the following transfer learning methods, described in Section ~\ref{sec:methods}:
\begin{itemize}
    \item \emph{Train on new data (baseline):} use only new data for training and start from a random initialization.
    \item \emph{Fine-tuning:} start from a source model and fine-tune it using new data.
    \item \emph{$L^2$-sp:} Fine-tuning with $L^2$ regularization for the model parameters. 
    \item \emph{Delta:} Fine-tuning with output matching regularization for the model parameters
\end{itemize}

These methods range from baselines to recently proposed approaches. 

Baseline networks have been trained for 35 epochs with a learning rate equal to $0.01$. Next, we used additional training for 15 epochs with a learning rate equal to $0.01$. We identified that $L^2$-sp and Delta are sensitive to choosing for fine-tuning learning rate and lowered it to $0.001$.

As we mentioned in Subsection~\ref{subsec:tl_conclusion}, domain adaptation concepts are used in our report to better estimate the representation learning models.

For each of the self-supervised approaches, we consider 5, 10 and 20 wells to get target models. Later, we use them to derive representations for the intervals in the target domain and cluster them using $\textbf{Formation\&class}$ target we described above for labelling. As a clustering technique, we apply Gaussian Mixture Model.

The exact settings of the experiments are given in Table ~\ref{tab:transfer_settings}.

\begin{table}[h!]
    \centering
    \caption{Experiment settings for transfer learning approach.}
    \begin{tabular}{ccc}
    \hline
    Method & Epoch number & Learning rate\\
    \hline
    Pretrained and Baseline & 35 &  0.01 for Siamese  \&  0.001 for other models \\ 
    Fine-tuning & 15 &  0.01 for Siamese  \&  0.001 for other models \\
    $L^2$ - sp & \multirow{2}{*}{15}& \multirow{2}{*}{0.001} \\ 
    Delta\\
    \hline
    \end{tabular}
    \label{tab:transfer_settings}
\end{table}

\begin{table}[h!]
    \centering
    \caption{Validation performance for four main models measured using four transfer learning scenarios. We consider a different number of wells in training data for each case. The best values are in \textbf{bold}. The second best values are \underline{underlined}.}
    \begin{tabular}{lcccccc}
    \hline
    \multirow{2}{*}{Training method} &
      \multicolumn{2}{c}{5 wells} &
      \multicolumn{2}{c}{10 wells} &
      \multicolumn{2}{c}{20 wells} \\
     & ARI & AMI  & ARI & AMI  & ARI & AMI  \\
    \hline
    \multicolumn{7}{c}{Triplet loss-based model} \\
    \hline
    Pretrained (Validation only)  & 0.26 & 0.44  & 0.26 & 0.44 & 0.26 & 0.44 \\
    Trained from scratch  & 0.23 & 0.43  & 0.20  & 0.41 & 0.30 & 0.41 \\
    Fine-tuning & 0.31 & 0.44 & 0.26 & 0.43 & 0.22 & 0.37\\
    $L^2$ - sp  & 0.24 & 0.41 & 0.28 & 0.46 & 0.30 & 0.43\\
    Delta & 0.39 & 0.45 & 0.21 & 0.44 & 0.30 & 0.43 \\
    \hline
    \multicolumn{7}{c}{Variational autoencoder} \\
    \hline
    Pretrained (Validation only)  & 0.27 & 0.47  & 0.27 & 0.47 & 0.27 & 0.47 \\
    Trained from scratch  & 0.32 & 0.41  & 0.25  & 0.40 & 0.30 & 0.50 \\
    Fine-tuning & 0.29 & 0.48 & 0.23 & 0.43 & 0.26 & 0.43\\
    $L^2$ - sp  & \underline{0.41} & 0.51 & 0.21 & 0.41 & 0.24 & 0.37\\
    Delta & 0.29 & 0.48 & 0.18 & 0.37 & 0.21 & 0.36 \\
    \hline
    \multicolumn{7}{c}{Autoregressive model} \\
    \hline
    Pretrained (Validation only)  & 0.34 & 0.49  & 0.34 & 0.49 & 0.34 & 0.49 \\
    Trained from scratch  & 0.35 & 0.49  & \underline{0.37}  & 0.50 & 0.34 & 0.48 \\
    Fine-tuning & \textbf{0.45} & \textbf{0.56} & \textbf{0.43} & 0.54 & \textbf{0.41} & 0.54\\
    $L^2$ - sp  & \textbf{0.45} & \textbf{0.56} & \underline{0.37} & 0.46 & \underline{0.40} & 0.52\\
    Delta & \underline{0.41}& \underline{0.54} & 0.35 & 0.48 & 0.38 & 0.52 \\
    \hline
    \multicolumn{7}{c}{Siamese model} \\
    \hline
    Pretrained (Validation only)  & 0.23 & 0.51  & 0.23 & 0.51 & 0.23 & 0.51 \\
    Trained from scratch  & 0.24 & 0.52  & 0.21  & 0.51 & 0.25 & 0.53 \\
    Fine-tuning & 0.24 & 0.53 & 0.27 & \textbf{0.56} & 0.26 & \underline{0.56}\\
    $L^2$ - sp  & 0.24 & 0.52 & 0.25 & \underline{0.55} & 0.27 & 0.55\\
    Delta & 0.22 & 0.51 & 0.24 & 0.53 & 0.27 & \textbf{0.57} \\
    \hline
    \end{tabular}
    \label{tab:validation_tl}
\end{table}

As shown in Table~\ref{tab:validation_tl}, autoregressive model shows the best performance for all the listed scenarios. 
The best approach for tuning differs for different models and a different number of training data available.
Also, we should mention that all the listed models encounter an overfitting problem when they are trained on more than 20 wells, as the quality degrades according to our metrics.
We expect that a proper selection of hyperparameters can fix these issues.

\begin{table}[ht!]
    \centering
    \caption{Reverse transfer from NZ data to Urenui for SiameseRNN. The model pretrained on the Urenui formation was transferred to 20 wells dataset from other New Zealand formations and then again validated on Urenui. The best values are in \textbf{bold}. The second best values are \underline{underlined}.}
    \begin{tabular}{cccc}
    \hline
    Training method & ARI & AMI & V-measure\\
    \hline
    Baseline  & \textbf{0.244} & \underline{0.499} & \underline{0.501} \\ 
    Fine-tuning     & 0.202 & 0.459 & 0.462\\  
    $L^2$ - sp    & \textbf{0.244} & 0.479 & 0.481\\
    Delta & \underline{0.240} & \textbf{0.500} & \textbf{0.502} \\
    \hline
    \end{tabular}
    \label{tab:reverse}
\end{table}

In this section, we investigated domain adaptation approaches to give us some insights into the ability of the models to generalize. More precisely, we can see that contrastive-based models (Triplet and SiameseRNN) do not benefit from transfer learning techniques until we start using at least 20 wells, whereas both Autoregressive and VAE models pretrained on source task and fine-tuned on target task shows dramatic improvements of the metric even for the case of just 5 wells.

\subsection{Saving source data patterns}
\label{subsec:rev_transfer}
In this part, we investigate models' ability to preserve source data patterns after transfer learning. While fine-tuning does not regulate at all the changes in the weights of the pretrained model, the $L^2$ - sp regularization is directly related to the similarity of the source and target models. In turn, Delta is aimed at the similarity of representations.

The results are presented in Table ~\ref{tab:reverse}.

\subsection{Classification validation}
\label{subsec:classification_results}

In order to verify the generalization for the obtained models, we will also consider the problem of embedding classification. Thus, we check how appropriate it is to use the representations obtained by self-supervised models for supervised learning.

In this Section, we will expand the study of the possibilities of the resulting embedding space by considering the classification problem. The first validation approach is based on the classification by tags from Section ~\ref{sec:results}. The second approach is based on the classification of the rock type label for Norwegian data. 

Our space of model in this experiment includes contrastive-based approach (Triplet), Variational Autoencoder and Autoregressive approaches. Results are presented in Table ~\ref{tab:geogr_target_clusterization_classification} , and the model settings for both classification experiments are given in Table \ref{tab:transfer_settings}.

\begin{table}[h!]
    \centering
    \caption{Accuracy for the classification for different true target values: geographical, formation and (formation, class) pairs. The best values are in \textbf{bold}. The second best values are \underline{underlined}.}
    \begin{tabular}{cccc}
    \hline
    & Baseline & Train from scratch & Fine-tuning\\
    \hline
    \multicolumn{4}{c}{Geographical} \\
    \hline
    Triplet  & 0.225 & 0.204 & \textbf{0.364} \\  
    VariationalAE      & \underline{0.307} & 0.303 & \underline{0.307}\\   
    Autoregressive     & 0.271 & 0.220 & 0.180\\   
    \hline
    \multicolumn{4}{c}{Formation} \\
    \hline
    Triplet  & 0.406 & 0.409 & \underline{0.652} \\
    VariationalAE      & 0.515 & 0.543 & \textbf{0.696}\\  
    Autoregressive     & 0.445 & 0.647 & 0.599\\  
    \hline
    \multicolumn{4}{c}{Class} \\
    \hline
    Triplet  & 0.082  & 0.151 & \underline{0.172} \\ 
    VariationalAE      & 0.127 & 0.140 & \textbf{0.174}\\  
    Autoregressive     & 0.098 & 0.114 & 0.108\\ 
    \hline
    \end{tabular}
    \label{tab:geogr_target_clusterization_classification}
\end{table}

\subsubsection{Rock type prediction}
Compared with other experiments, the rock type is not tied to either the formation or the class of the well. Our task is to test the previously obtained representations' flexibility when applied to indirect problems, such as predicting the type of rock.

The Norwegian data contains 12 types of rocks described in ~\cite{romanenkova2022similarity} and distributed between 12 wells. Results are presented in Table~\ref{tab:rock_type}.
We see, that our representations can be used for rock type prediction.

\begin{table}
    \centering
    \caption{Accuracy for the classification of rock types for Norwegian data. The best values are in \textbf{bold}. The second best values are \underline{underlined}.}
    \begin{tabular}{cccc}

    \hline
    & Baseline & Train from scratch & Fine-tuning\\
    \hline
    Triplet  & 0.403 & 0.337 & 0.407 \\ 
    VariationalAE      & \underline{0.483} & 0.399 & \textbf{0.484}\\  
    Autoregressive    & 0.479 & 0.471 & 0.418\\  
    \hline
    \end{tabular}
    \label{tab:rock_type}
\end{table}

\section{Conclusion}
\label{sec:conclusion}

We considered the problem of representation learning for well logging data.
As we often have little labelled data, we consider approaches that use only unlabeled data from representation training.
Such methods are called self-supervised. 
After making necessary modifications, we compare the main paradigms for self-supervised learning: generative (autoregression) and contrastive.

Among the considered methods, the best results were obtained for an autoregression model and close results for a variational autoencoder. 
It is different from computer vision, where contrastive approaches show better performance. 
We expect this effect to happen for two reasons: a more significant amount of data available in computer vision and more structured data compared to time series.
Our representations show their success in both clustering and classification scenarios.

In addition, to test our models, we transferred the obtained representations to both the direct task of embedding classification and the indirect task of predicting the rock type.
The variational autoencoder shows high results in all cases and is the best model for generalizing representations.

For transfer learning, fine-tuning is the most useful strategy. 
For small datasets, it is worth using fine-tuning or $L^2-sp$; both methods are superior to learning from scratch. However, the fine-tuning model forgets the patterns obtained from the source data. 
At the same time, the other considered method, Delta, does not give a noticeable advantage at the initial size of the training sample but allows for the accumulation of patterns on different data better than $L^2-sp$ and even more efficiently than the base model. 
When the training sample increases, fine-tuning is ahead of other methods and allows to get results similar to learning from scratch on the entire formation, only in a much shorter time.
In general, we provide evidence, that our representations can be reused for other data without much additional data and training, making them universal across oil fields and formations.


\bibliographystyle{cas-model2-names}
\bibliography{bibliography} 

\appendix

\section{Transfer learning}
\label{sec:transfer_app}

This section provides detailed transfer learning training and a selection of hyperparameters for the SiameseRNN.

\subsection{Transfer learning for Siamese model}
\label{sec:transfer_simple}

We split the test data to two parts: 10 wells are used for test, while the remaining ones were used for fine-tuning of transferred models.
To identify the efficiency of transfer we vary the number of wells used for training and report the dynamic of the model quality, as more data fall into the training dataset. Training from scratch took 200 epochs and the rest of the methods - took 25; we also used a sliding learning rate from $0.01$.

\begin{figure}
    \centering
    \resizebox{\columnwidth}{!}{%
    \includegraphics[width=1.3\textwidth]{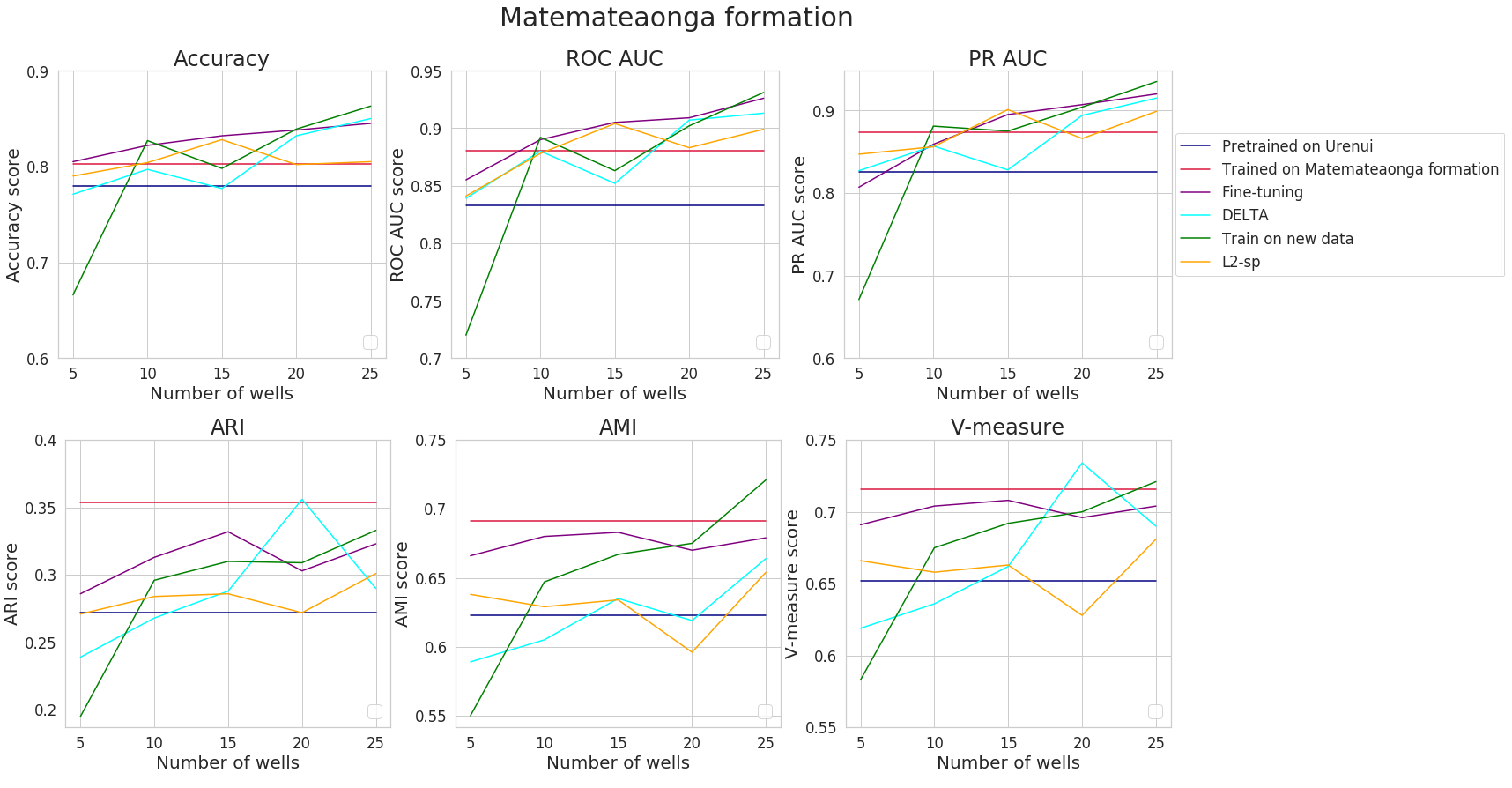}%
    }
    \caption{Transfer Learning on Matemateaonga formation.}
    \label{fig:transfer_learning_Matemateaonga}
\end{figure}

\begin{figure}
    \centering
    \resizebox{\columnwidth}{!}{%
    \includegraphics[width=1.3\textwidth]{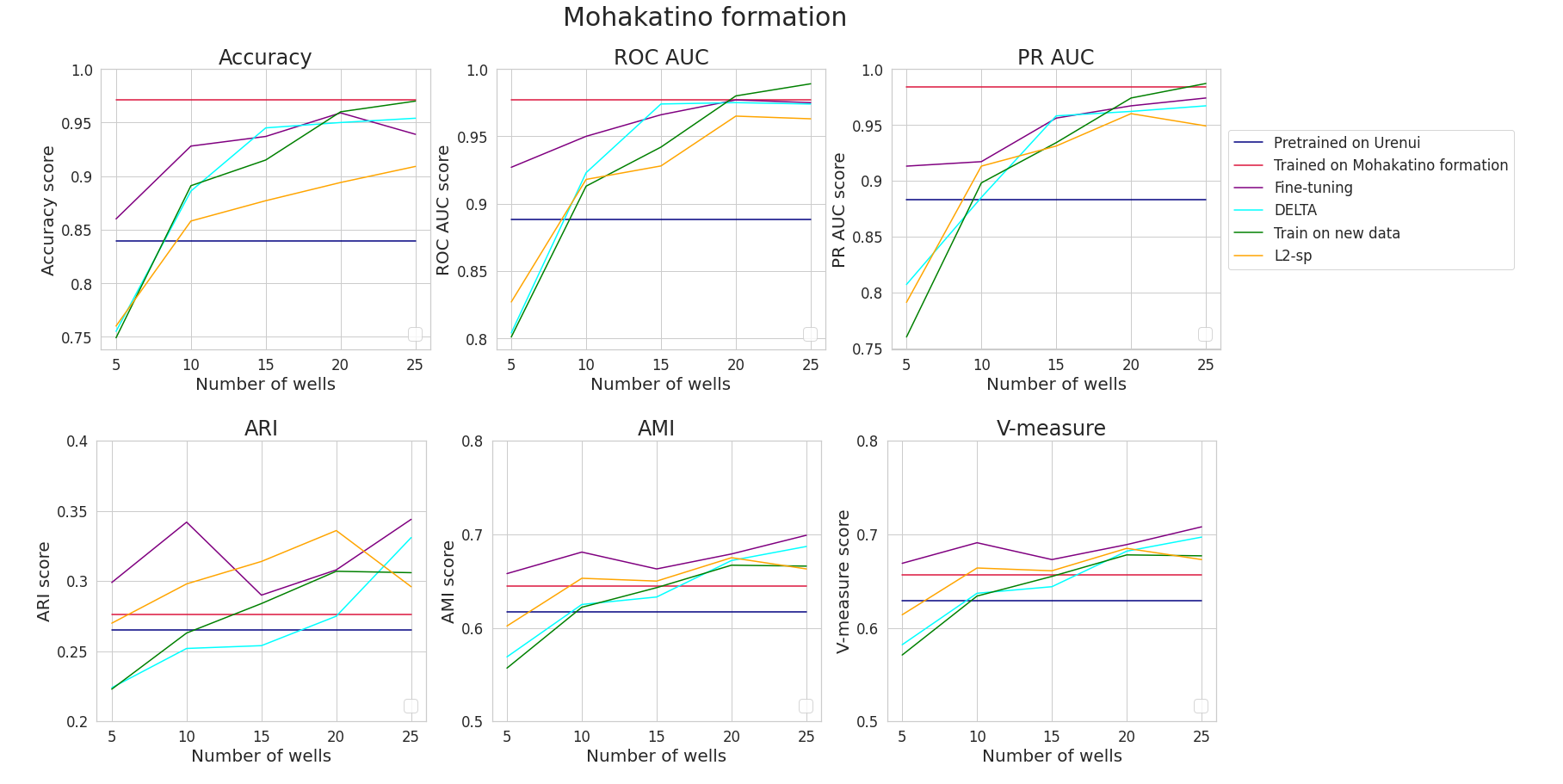}%
    }
    \caption{Transfer Learning on Mohakatino formation.}
    \label{fig:transfer_learning_mohakatino}
\end{figure}

\begin{figure}
    \centering
    \resizebox{\columnwidth}{!}{%
    \includegraphics[width=1.3\textwidth]{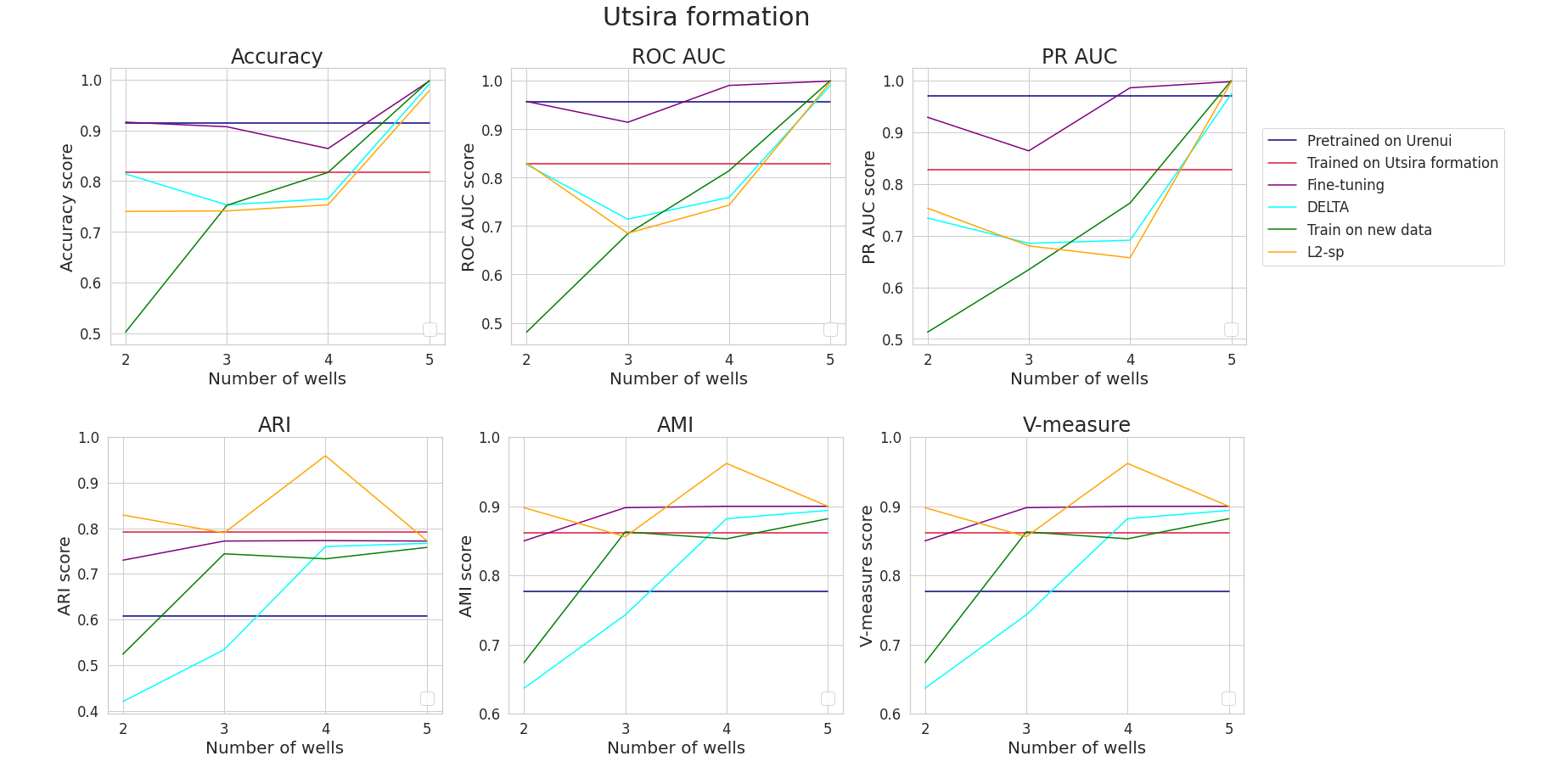}%
    }
    \caption{Transfer Learning on Utsira formation.}
    \label{fig:transfer_learning_utsira}
\end{figure}

We present results for different target formations in Tables ~\ref{tab:transfer_learning_Matemateaonga}, ~\ref{tab:transfer_learning_mohakatino} and ~\ref{tab:transfer_learning_utsira}. 
Visualization of the results can be seen in Figures ~\ref{fig:transfer_learning_Matemateaonga}, ~\ref{fig:transfer_learning_mohakatino} and ~\ref{fig:transfer_learning_utsira}. 
The lower horizontal line shows the initial quality of the metrics of the pre-trained model, while the upper one shows the quality when training from scratch on the entire new formation, with the exception of test wells. Thus, the parts of the graphs above the upper horizontal line correspond to the choice in favor of transfer learning instead of learning from scratch.

\begin{table}[!h]
\centering
\caption{Transfer Learning results on Matemateaonga formation. The best values are in \textbf{bold}. The second best values are \underline{underlined}.}
\resizebox{\columnwidth}{!}{%
\footnotesize
\begin{tabular}{cccccccc} 
\hline
Number of wells & Method & Accuracy & ROC AUC & PR AUC & ARI & AMI & V-measure \\ \hline 
All dataset, with the & Pretrained on Urenui & 0.780 & 0.833 & 0.826 & 0.272 & 0.623 & 0.652\\ 
exception of test wells & Trained on Matemateaonga & 0.803 & 0.880 & 0.874 & \textbf{0.354} & \underline{0.691} & 0.716\\ \hline 
\multirow{4}{*}{5} & Fine-Tuning & 0.805 & 0.855 & 0.807 & 0.286 & 0.666 & 0.691\\ 
 & L2-sp & 0.790 & 0.841 & 0.847 & 0.271 & 0.638 & 0.666\\ 
 & Delta & 0.771 & 0.839 & 0.827 & 0.239 & 0.589 & 0.619\\ 
 & Train on new data & 0.666 & 0.720 & 0.671 & 0.195 & 0.550 & 0.583\\ \hline
 
\multirow{4}{*}{10} & Fine-Tuning & 0.822 & 0.890 & 0.859 & 0.313 & 0.680 & 0.704\\ 
 & L2-sp & 0.804 & 0.878 & 0.856 & 0.284 & 0.629 & 0.658\\ 
 & Delta & 0.797 & 0.880 & 0.857 & 0.268 & 0.605 & 0.636\\ 
 & Train on new data & 0.827 & 0.892 & 0.881 & 0.296 & 0.647 & 0.675\\ \hline
 
\multirow{4}{*}{15} & Fine-Tuning & 0.832 & 0.905 & 0.895 & 0.332 & 0.683 & 0.708\\ 
 & L2-sp & 0.828 & 0.904 & 0.901 & 0.286 & 0.634 & 0.663\\ 
 & Delta & 0.777 & 0.852 & 0.828 & 0.288 & 0.635 & 0.662\\ 
 & Train on new data & 0.798 & 0.863 & 0.875 & 0.310 & 0.667 & 0.692\\ \hline
 
\multirow{4}{*}{20} & Fine-Tuning & 0.838 & 0.909 & 0.907 & 0.303 & 0.670 & 0.696\\ 
 & L2-sp & 0.802 & 0.883 & 0.866 & 0.272 & 0.596 & 0.628\\ 
 & Delta & 0.832 & 0.907 & 0.894 & 0.356 & 0.619 & \textbf{0.734}\\ 
 & Train on new data & 0.839 & 0.902 & 0.904 & 0.309 & 0.675 & 0.700\\ \hline
 
\multirow{4}{*}{25} & Fine-Tuning & 0.845 & \underline{0.926} & \underline{0.920} & 0.323 & 0.679 & 0.704\\ 
 & L2-sp & 0.805 & 0.899 & 0.899 & 0.301 & 0.654 & 0.681\\ 
 & Delta & \underline{0.850} & 0.913 & 0.915 & 0.290 & 0.664 & 0.690\\ 
 & Train on new data & \textbf{0.863} & \textbf{0.931} & \textbf{0.935} & \underline{0.333} & \textbf{0.721} & \underline{0.721}\\ \hline
\end{tabular}%
}
    \label{tab:transfer_learning_Matemateaonga}
\end{table}

\begin{table}[!h]
\centering
\caption{Transfer Learning results on Mohakatino formation. The best values are in \textbf{bold}. The second best values are \underline{underlined}.}
\resizebox{\columnwidth}{!}{%
\footnotesize
\begin{tabular}{cccccccc} 
\hline
Number of wells & Method & Accuracy & ROC AUC & PR AUC & ARI & AMI & V-measure \\ \hline 
All dataset, with the & Pretrained on Urenui & 0.839 & 0.888 & 0.883 & 0.265 & 0.617 & 0.629\\ 
exception of test wells & Trained on Mohakatino & \textbf{0.971} & \underline{0.977} & \underline{0.984} & 0.276 & 0.645 & 0.657\\ \hline 
\multirow{4}{*}{5} & Fine-Tuning & 0.860 & 0.927 & 0.913 & 0.299 & 0.658 & 0.669\\ 
 & L2-sp & 0.760 & 0.827 & 0.791 & 0.270 & 0.602 & 0.614\\ 
 & Delta & 0.755 & 0.804 & 0.807 & 0.224 & 0.569 & 0.582\\ 
 & Train on new data & 0.749 & 0.801 & 0.760 & 0.223 & 0.557 & 0.571\\ \hline
 
\multirow{4}{*}{10} & Fine-Tuning & 0.928 & 0.950 & 0.917 & \underline{0.342} & 0.681 & 0.691\\ 
 & L2-sp & 0.858 & 0.918 & 0.913 & 0.298 & 0.653 & 0.664\\ 
 & Delta & 0.886 & 0.923 & 0.885 & 0.252 & 0.625 & 0.637\\ 
 & Train on new data & 0.891 & 0.913 & 0.898 & 0.263 & 0.622 & 0.634\\ \hline
 
\multirow{4}{*}{15} & Fine-Tuning & 0.937 & 0.966 & 0.956 & 0.290 & 0.663 & 0.673\\ 
 & L2-sp & 0.877 & 0.928 & 0.931 & 0.314 & 0.650 & 0.661\\ 
 & Delta & 0.945 & 0.974 & 0.958 & 0.254 & 0.633 & 0.644\\ 
 & Train on new data & 0.915 & 0.942 & 0.934 & 0.284 & 0.643 & 0.655\\ \hline
 
\multirow{4}{*}{20} & Fine-Tuning & 0.959 & \underline{0.977} & 0.967 & 0.308 & 0.679 & 0.689\\ 
 & L2-sp & 0.894 & 0.965 & 0.960 & 0.336 & 0.675 & 0.685\\ 
 & Delta & 0.950 & 0.975 & 0.962 & 0.275 & 0.672 & 0.682\\ 
 & Train on new data & 0.960 & 0.980 & 0.974 & 0.307 & 0.667 & 0.678\\ \hline
 
\multirow{4}{*}{25} & Fine-Tuning & 0.939 & 0.975 & 0.974 & \textbf{0.344} & \textbf{0.699} & \textbf{0.708}\\ 
 & L2-sp & 0.909 & 0.963 & 0.949 & 0.296 & 0.663 & 0.673\\ 
 & Delta & 0.954 & 0.974 & 0.967 & 0.331 & \underline{0.687} & \underline{0.697}\\ 
 & Train on new data & \underline{0.970} & \textbf{0.989} & \textbf{0.987} & 0.306 & 0.666 & 0.677\\ \hline
\end{tabular}%
}
    \label{tab:transfer_learning_mohakatino}
\end{table}

\begin{table}[!h]
\centering
\caption{Transfer Learning results on Utsira formation. The best values are in \textbf{bold}. The second best values are \underline{underlined}.}
\resizebox{\columnwidth}{!}{%
\footnotesize
\begin{tabular}{cccccccc} 
\hline
Number of wells & Method & Accuracy & ROC AUC & PR AUC & ARI & AMI & V-measure \\ \hline 
All dataset, with the & Pretrained on Urenui & 0.914 & 0.957 & 0.970 & 0.607 & 0.777 & 0.777\\ 
exception of test wells & Trained on Utsira & 0.818 & 0.828 & 0.828 & 0.792 & 0.861 & 0.861\\ \hline 
\multirow{4}{*}{2} & Fine-Tuning & 0.916 & 0.957 & 0.929 & 0.730 & 0.850 & 0.850\\
                    &L2-sp & 0.740 & 0.830 & 0.753 & \underline{0.829} & 0.898 & 0.898\\
                    &Delta & 0.814 & 0.827 & 0.734 & 0.420 & 0.637 & 0.637\\
                    &Train on new data & 0.502 & 0.481 & 0.513 & 0.524 & 0.674 & 0.674\\ \hline
 
\multirow{4}{*}{3} & Fine-Tuning & 0.907 & 0.914 & 0.864 & 0.772 & 0.898 & 0.898\\
                    & L2-sp & 0.741 & 0.685 & 0.680 & 0.790 & 0.856 & 0.856\\
                    & Delta & 0.753 & 0.714 & 0.685 & 0.534 & 0.743 & 0.743\\
                    & Train on new data & 0.752 & 0.684 & 0.634 & 0.744 & 0.863 & 0.863\\ \hline
 
\multirow{4}{*}{4} & Fine-Tuning & 0.864 & 0.990 & 0.986 & 0.773 & \underline{0.900} & \underline{0.900}\\
                    & L2-sp & 0.753 & 0.743 & 0.657 & \textbf{0.959} & \textbf{0.962} & \textbf{0.962}\\
                    & Delta & 0.765 & 0.759 & 0.691 & 0.760 & 0.882 & 0.882\\
                    & Train on new data & 0.817 & 0.814 & 0.763 & 0.733 & 0.853 & 0.853\\ \hline
 
\multirow{4}{*}{5} & Fine-Tuning & \underline{0.997} & \underline{0.999} & \underline{0.998} & 0.772 & \underline{0.900} & \underline{0.900}\\
                    & L2-sp & 0.978 & 0.997 & \underline{0.998} & 0.773 & \underline{0.900} & \underline{0.900}\\
                    & Delta & 0.991 & 0.990 & 0.974 & 0.767 & 0.894 & 0.894\\
                    & Train on new data & \textbf{0.998} & \textbf{1.000} & \textbf{1.000} & 0.758 & 0.882 & 0.882\\ \hline
\end{tabular}%
}
    \label{tab:transfer_learning_utsira}
\end{table}

\subsection{One-two wells transfer learning for close well linking problem}

To see how transfer algorithms work at the initial stages of training, we conducted a series of transfer training experiments that use only one and two wells of a new formation for the close well linking problem.
 describe how we train similarity model using data only from one well.

The results are shown in Table~\ref{tab:transfer_learning_one_two}. 

Fine-tuning allows good transfer even using data from one or two wells from a new formation.
Fine-tuning is the best strategy for transfer learning on large formations, such as Matemateaonga (45 wells) and Mohakatino (41 wells), while $L^2-sp$ shows the best results for Utsira (12 wells) when training on a single well. 

\begin{table}[!h]
\centering
\caption{Transfer Learning results for the one-two close well linking problem. The best values are in \textbf{bold}. The second best values are \underline{underlined}.}
\resizebox{\columnwidth}{!}{%
\footnotesize
\begin{tabular}{cccccccc}\hline 
Method & Number of wells & Accuracy & ROC AUC & PR AUC & ARI & AMI & V-measure \\ \hline 
\multicolumn{8}{c}{\textbf{Matemateaonga formation}} \\\hline 
\multirow{2}{*}{$Fine-tuning$} & 1 & 0.758  & 0.835  & 0.708  & 0.377  & 0.597  & 0.598   \\ 
& 2 & \textbf{0.778   } & 0.835    & \textbf{0.720  } &  \textbf{0.577   } & 0.730    & 0.730   \\ \hline 
\multirow{2}{*}{$L^2-sp$} &1 &  0.673    & 0.724    & 0.528   & 0.336    & 0.631    & 0.635    \\ 
& 2 & 0.717    & 0.818    & 0.644    & 0.452    & \textbf{0.732   } & \textbf{0.736   }\\ \hline 
\multirow{2}{*}{$Delta$} & 1 & 0.672    & 0.809   & 0.649    & 0.219    & 0.559    & 0.593   \\ 
& 2 & 0.701    & \textbf{0.841   } & 0.699    & 0.252    & 0.597    & 0.629   \\ \hline 
\multicolumn{8}{c}{\textbf{Mohakatino formation}} \\\hline 
\multirow{2}{*}{$Fine-tuning$} &1 &  0.780    & 0.815    & 0.541    & \textbf{0.476   } & 0.651    & 0.651   \\ 
& 2 &  \textbf{0.803   } & 0.831   & \textbf{0.637  } & 0.475    & \textbf{0.711   } & \textbf{0.711   } \\ \hline 
\multirow{2}{*}{$L^2-sp$} &1 &  0.741   & 0.715    & 0.436   & 0.294    & 0.614    & 0.618   \\ 
&2 & 0.782    & \textbf{0.857   } & 0.572    & 0.344    & 0.655   & 0.657   \\\hline 
\multirow{2}{*}{$Delta$} &1 &  0.768    & 0.827    & 0.540   & 0.269   & 0.602   & 0.614  \\ 
& 2 & 0.780    & 0.853    & 0.540    & 0.278    & 0.623   & 0.635  \\\hline  
\multicolumn{8}{c}{\textbf{Utsira formation}} \\\hline 
\multirow{2}{*}{$Fine-tuning$} &1 &  0.515     & 0.554   & 0.851    & 0.443    & 0.646    & 0.646    \\ 
&2 &  \textbf{0.754    } & 0.550    & \textbf{0.854   } & 0.629    & 0.854    & 0.854  \\\hline 
\multirow{2}{*}{$L^2-sp$} &1 &  0.642     & 0.758    & 0.737    & 0.810    & \textbf{0.869   } & \textbf{0.870   }\\ 
& 2 &  0.717     & 0.818    & 0.644    & 0.728    & 0.818    & 0.818   \\ \hline
\multirow{2}{*}{$Delta$} &1 &  0.479     & 0.770    & 0.762   & \textbf{0.812   } & 0.850    & 0.850    \\ 
&2 &  0.669     & \textbf{0.820   } & 0.734    & 0.559    & 0.681    & 0.681    \\ \hline
\end{tabular}%
}
    \label{tab:transfer_learning_one_two}
\end{table}

\subsection{Selection of hyperparameters for transfer learning}
\label{sec:hyperparameters_self_supervised}

\begin{figure}[!ht]
    \centering
    \includegraphics[width=1\linewidth]{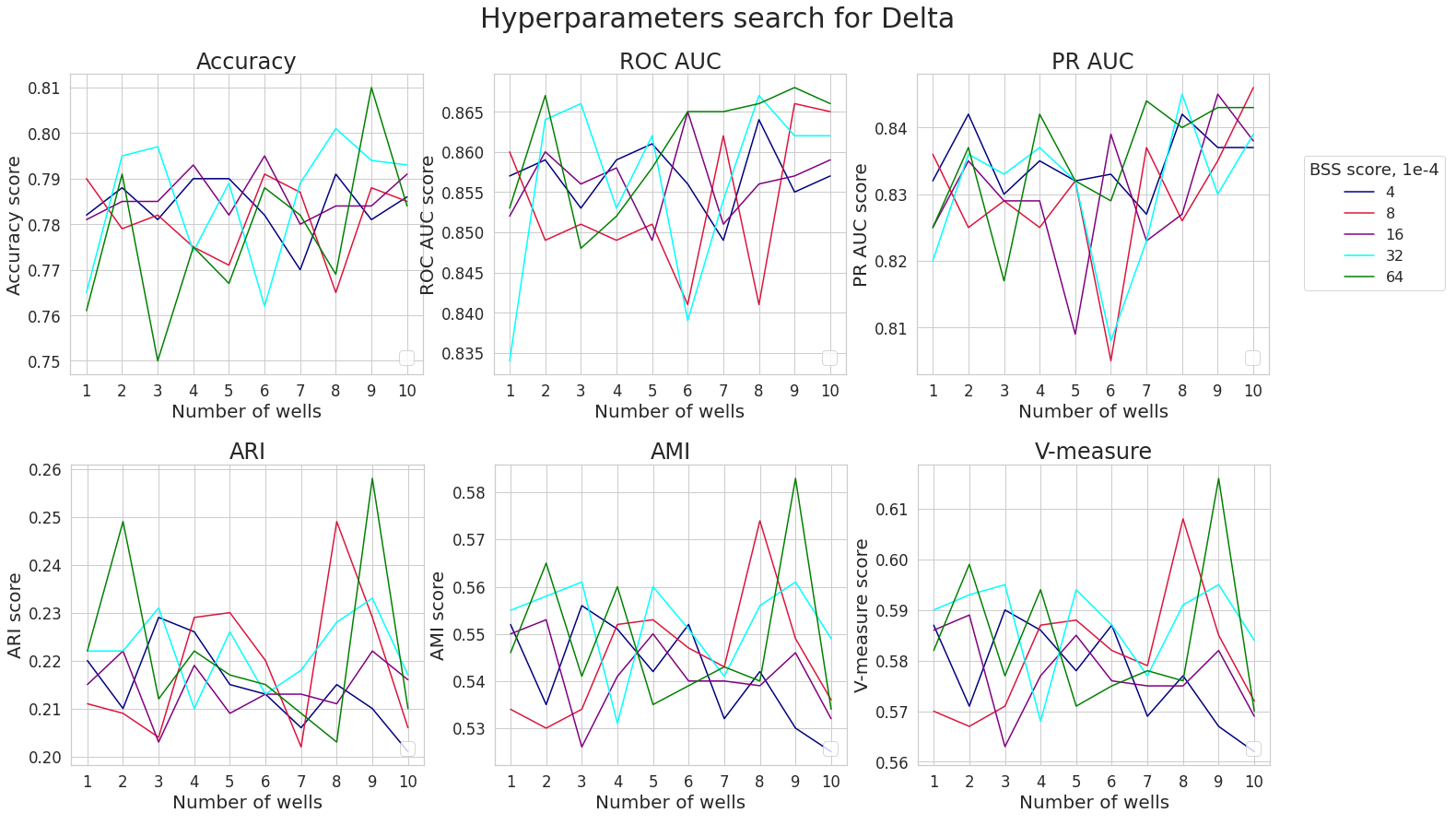}
    \caption{Delta hyperparameters search.}
    \label{fig:fm}
\end{figure}

Here we present the results of the selection of hyperparameters for regularizer $Delta + BSS$.
In Figure ~\ref{fig:fm} we present how variation of BSS and Delta hyperparameters lead to different performances for different hyperparameters values.

As we can see, all launches with different BSS parameters lie approximately in the same interval. Moreover, additional training leads to an increase in quality compared to training from scratch (Section ~\ref{sec:results}). Therefore, we conclude that the problem of negative transfer is not relevant to our data, so we do not use this add-in in our final experiments.
As for Delta, choosing too small or too large parameters leads to lower metric scores - in the first case, regilarization practically does not contribute to the total loss, while at large values, binary entropy is poorly optimized.

\subsubsection{Close well linking value}

Here we present an approach to close well linking parameter selection. A detailed description of the method can be found in the Section ~\ref{sec:methods}. 

Outside of the depth interval set by this parameter, samples from one well are labeled 0, as are samples from different wells. Thus, we expect that when the close well linking parameter is increased, samples from larger depth intervals will be recognized as similar. For fully recognizable wells, the close well linking task is the same as the well linking task.

So, we investigated the change in the quality of clustering metrics depending on the value of the parameter and compared it with the depths of wells in the training sample for each formation separately. 

\begin{figure}[!ht]
    \centering
    \includegraphics[width=0.5\textwidth]{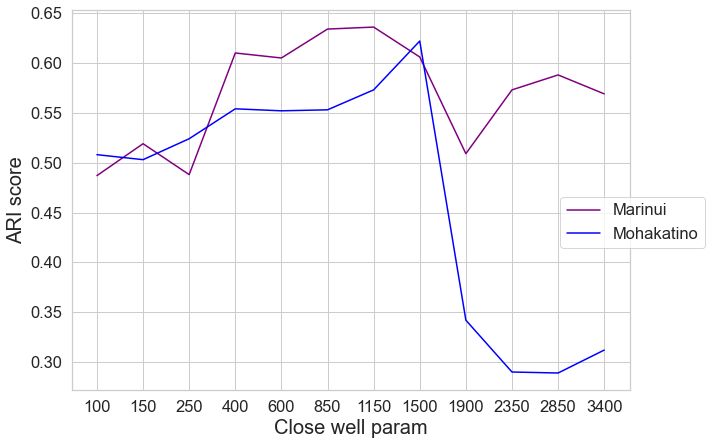}
    \caption{ARI metric score for close well linking hyperparameter search.}
    \label{fig:cwl_hype_s_m_m}
\end{figure}

\begin{figure}[!ht]
    \centering
    \resizebox{\columnwidth}{!}{%
    \includegraphics[width=1.25\linewidth]{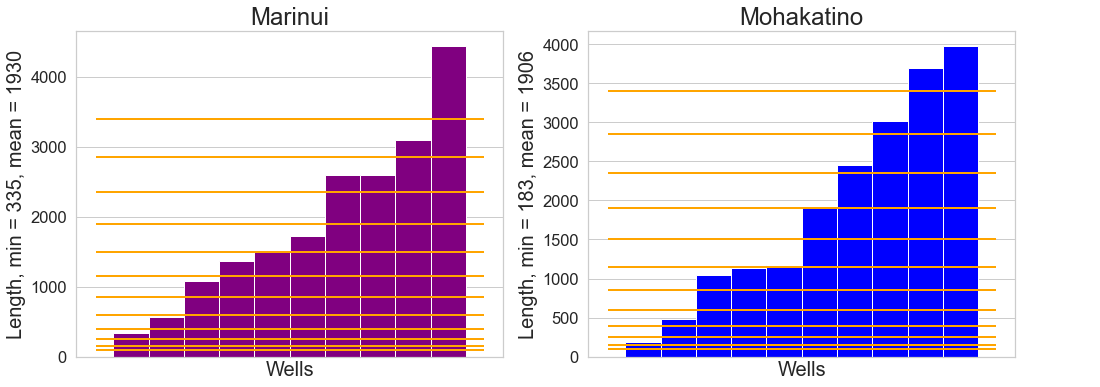}%
    }
    \caption{Training datasets for close well linking hyperparameter search. Orange lines - the length of the interval at which similar samples are recognized, set by the close well linking parameter. For wells located below the orange line set by the corresponding parameter value, the close well linking task coincides with the well linking task.}
    \label{fig:hype_s_train_short}
\end{figure}

\begin{figure}[!ht]
    \centering
    \resizebox{\columnwidth}{!}{%
    \includegraphics[width=1.25\linewidth]{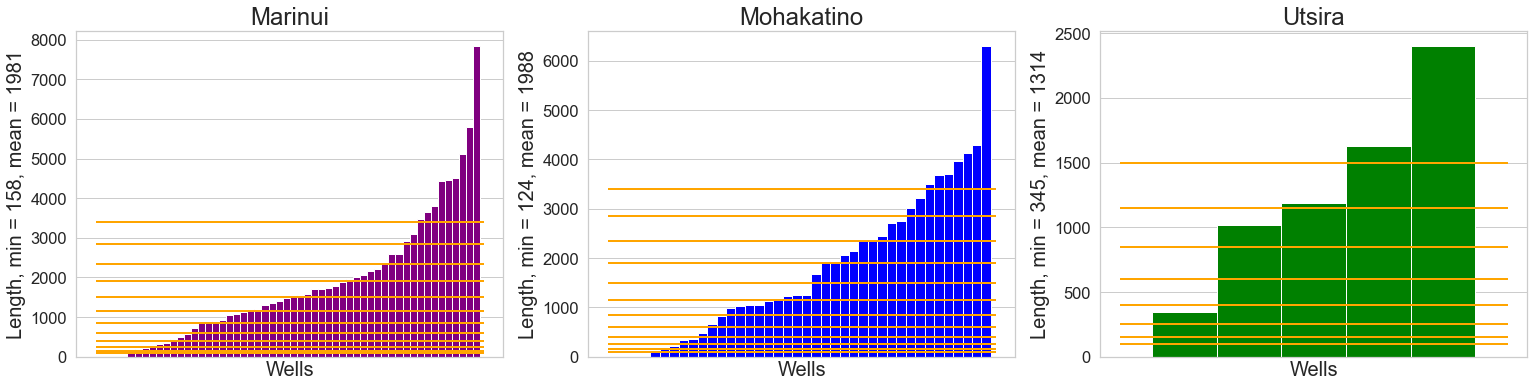}%
    }
    \caption{Datasets analysis for close well linking problem. Orange lines - the length of the interval, set by the close well linking parameter. For wells located below the orange line set by the corresponding parameter value, the close well linking task coincides with the well linking task.}
    \label{fig:hype_s_all}
\end{figure}

The corresponding results are shown in Figure ~\ref{fig:cwl_hype_s_m_m}. The visualization of the training datasets is represented by the Figure ~\ref{fig:hype_s_train_short}, and for all dataset on Figure~\ref{fig:hype_s_all}.

Thus, the best metric score are achieved at close linking parameter values slightly smaller than the average length of wells in the formation - at such values, the close linking problem has large differences with well linking, allowing the model to notice more local dependencies.

\subsection{Conclusions}

Fine-tuning remains the best strategy for our task and shows results similar to learning from scratch and even better. At the same time, Delta does not give a noticeable advantage at the initial size of the training sample, and when the training sample is increased, it ceases to compete with other methods. Experiments on transfer learning from formations with a large number of wells to small formations have shown that even at the initial stages of training, fine-tuning and $L^2-sp$ regularization perform better than training the Siamese network on the entire dataset of the new formation. However, $L^2-sp$ has higher starting metrics and should be used as a priority.

For the close well linking problem, we investigated the dependence of the Siamese network behavior on the parameter value. The highest values of the metrics are achieved when the parameter is close to the average depth of the wells of the formation.

\end{document}